\documentclass[runningheads]{llncs}
\usepackage{amsmath}
\usepackage[T1]{fontenc}
\usepackage[hidelinks, colorlinks=false, linkcolor=red, citecolor=blue, urlcolor=blue]{hyperref}
\usepackage{graphicx,verbatim}
\usepackage{multirow}
\usepackage{makecell}
\usepackage{arydshln} 

\usepackage[T1]{fontenc}
\usepackage{graphicx,verbatim}
\begin{document}
\title{FIND-Net – Fourier-Integrated Network with Dictionary Kernels for Metal Artifact Reduction}
\titlerunning{FIND-Net – Fourier-Integrated Network with Dictionary Kernels for MAR}

\author{Farid Tasharofi\inst{1}\orcidID{0009-0006-1822-7889} \and
Fuxin Fan\inst{1} \and
Melika Qahqaie\inst{1} \and
\\ Mareike Thies\inst{1} \and
Andreas Maier\inst{1}}

\authorrunning{F. Tasharofi et al.}
%
\institute{Pattern Recognition Lab, Friedrich-Alexander-Universität Erlangen, Germany 
\email{farid.tasharofi@fau.de}}


\maketitle              
\begin{abstract}
Metal artifacts, caused by high-density metallic implants in computed tomography (CT) imaging, severely degrade image quality, complicating diagnosis and treatment planning. While existing deep learning algorithms have achieved notable success in Metal Artifact Reduction (MAR), they often struggle to suppress artifacts while preserving structural details. To address this challenge, we propose FIND-Net (\textbf{F}ourier-\textbf{I}ntegrated \textbf{N}etwork with \textbf{D}ictionary Kernels), a novel MAR framework that integrates frequency and spatial domain processing to achieve superior artifact suppression and structural preservation. 
FIND-Net incorporates Fast Fourier Convolution (FFC) layers and trainable Gaussian filtering, treating MAR as a hybrid task operating in both spatial and frequency domains. This approach enhances global contextual understanding and frequency selectivity, effectively reducing artifacts while maintaining anatomical structures. Experiments on synthetic datasets show that FIND-Net achieves statistically significant improvements over state-of-the-art MAR methods, with a 3.07\% MAE reduction, 0.18\% SSIM increase, and 0.90\% PSNR improvement, confirming robustness across varying artifact complexities. Furthermore, evaluations on real-world clinical CT scans confirm FIND-Net’s ability to minimize modifications to clean anatomical regions while effectively suppressing metal-induced distortions. These findings highlight FIND-Net’s potential for advancing MAR performance, offering superior structural preservation and improved clinical applicability. Code is available at \href{https://github.com/Farid-Tasharofi/FIND-Net}{this link}~\footnote{Code is available at \url{https://github.com/Farid-Tasharofi/FIND-Net}}.



\end{abstract}

\section{Introduction}
Computed tomography (CT) plays a crucial role in non-invasive diagnostics; however, metallic implants introduce severe artifacts, such as streaks and star-shaped distortions, arising from beam hardening, photon starvation, and scattering \cite{gjesteby2016metal,wang2023oscnet,anhaus2024new}. These artifacts obscure anatomical structures and degrade accuracy.

Traditional metal artifact reduction (MAR) methods, such as linear interpolation (LI), are computationally efficient but often introduce secondary artifacts, compromising structural fidelity \cite{anhaus2024new,kalender1987reduction}. Deep learning-based MAR approaches, particularly convolutional neural networks (CNNs), have demonstrated superior artifact suppression~\cite{wang2023oscnet,Lyu_2020,yu2020deep,wang2021dicdnet}. However, their limited receptive fields hinder the effective handling of large-scale, globally distributed distortions. The inability to effectively capture long-range dependencies in both image and sinogram domains remains a critical limitation in CNN methods \cite{richter2021should}. 

CNNs predominantly rely on small-kernel convolutions (e.g., $3 \times 3$ in ResNet), which emphasize local features, making global artifact suppression challenging \cite{richter2021should,chi2020fast}. Sinogram-domain processing, while aiming to adhere to imaging system constraints, often introduces secondary distortions or localized oversmoothing. Conversely, image reconstruction from metal-corrupted sinograms propagates metal-induced artifacts throughout the image, leading to globally distributed distortions~\cite{yu2020deep}. Frequency-domain techniques offer a promising avenue for global feature extraction \cite{chi2020fast}; however, their integration with deep learning remains limited, as most MAR methods primarily operate in the spatial domain and struggle with long-range dependencies.

To address these challenges, we propose FIND-Net (Fourier-Integrated Network with Dictionary Kernels), a novel MAR model that integrates Fast Fourier Convolution (FFC) with trainable Gaussian filtering for improved artifact suppression. By incorporating both spatial and frequency-domain processing, FIND-Net captures long-range dependencies while preserving structural details, reducing metal-induced distortions without oversmoothing anatomical structures.

\section{Methodology}
\subsubsection{Problem Formulation}
FIND-Net extends the convolutional dictionary model from DICDNet~\cite{wang2021dicdnet}, which decomposes a metal-corrupted CT image \( \boldsymbol{Y} \) into an artifact-free component \( \boldsymbol{X} \) and an artifact term \( \boldsymbol{A} \):

\begin{equation} 
\boldsymbol{I} \odot \boldsymbol{Y} = \boldsymbol{I} \odot \boldsymbol{X} + \boldsymbol{I} \odot \boldsymbol{A},
\label{eq:dicdnet1}
\end{equation}
where \( \boldsymbol{I} \) is a non-metal mask, and \( \odot \) denotes point-wise multiplication. The artifact term is modeled as:

\begin{equation}
\boldsymbol{A} = \sum_{n=1}^{N} \boldsymbol{K}_n \otimes \boldsymbol{M}_n.
\label{eq:dicdnet2}
\end{equation}

Here, \( \mathcal{K} = [\boldsymbol{K}_1, \dots, \boldsymbol{K}_N] \) are dictionary kernels capturing metal artifact patterns, and \( \mathcal{M} = [\boldsymbol{M}_1, \dots, \boldsymbol{M}_N] \) are the corresponding feature maps. The operator \( \otimes \) represents a 2D convolution. This decomposition encodes metal artifact structures using localized patterns and spatial priors, facilitating interpretability in MAR~\cite{wang2021dicdnet}. However, estimating both \( \boldsymbol{M} \) and \( \boldsymbol{X} \) is inherently ill-posed, requiring an optimization framework to ensure stability and effective MAR.

\subsubsection{Optimization Framework:}
Combining Equations \ref{eq:dicdnet1} and \ref{eq:dicdnet2}, we can reformulate MAR as an inverse problem requiring constraints for stable optimization. However, the problem is ill-posed due to non-uniqueness, instability, and an underdetermined nature, requiring additional constraints for a meaningful solution. To address this, DICDNet~\cite{wang2021dicdnet} introduce a regularized optimization framework:~\begin{equation} 
\min _{\mathcal{M}, {\boldsymbol{X}}} \left\| {\boldsymbol{I}} \odot \left( {\boldsymbol{Y}} - {\boldsymbol{X}} - \mathcal{A} \right) \right\|_{F}^{2} 
+ \lambda_{1} P_{1}(\mathcal{M}) + \lambda_{2} P_{2}({\boldsymbol{X}}),
\end{equation}
where \( P_{1}(\mathcal{M}) \) and \( P_{2}({\boldsymbol{X}}) \) are prior terms that constrain \( \mathcal{M} \) and \( \boldsymbol{X} \) to regularize the solution and address ill-posedness. The parameters \( \lambda_{1} \) and \( \lambda_{2} \) are weighting factors that control the influence of the regularization terms on the optimization process. The iterative updates are formulated as:

\begin{equation}
\mathcal{M}^{(s)} = \operatorname{prox}_{\lambda _{1} \eta_1}\left(\mathcal{M}^{(s-1)} - \eta_1 \mathcal{K} \otimes^T \left(\boldsymbol{I} \odot\left(\mathcal{A}^{(s-1)} + \boldsymbol{X}^{(s-1)} - \boldsymbol{Y}\right)\right)\right),
\end{equation}

\begin{equation}
\boldsymbol{X}^{(s)} = \operatorname{prox}_{\lambda _{2} \eta_2}\left( ( 1 - \eta _ { 2 } \boldsymbol { I } ) \odot \boldsymbol { X } ^ { ( s - 1 ) } + \eta _ { 2 } \boldsymbol { I } \odot ( \boldsymbol { Y } - \mathcal{A}^{(s)} ) \right),
\end{equation}
where \( \mathcal{A}^{(s)} = \mathcal{K} \otimes \mathcal{M}^{(s)} \). The proximal operators ensure stability while enforcing constraints. As in DICDNet~\cite{wang2021dicdnet}, \( X^{(0)} \) is initialized using LI to provide a reasonable starting point for iterative refinement. DICDNet~\cite{wang2021dicdnet} employs ResNet~\cite{he2016deep} as a proximal operator to iteratively refine both the estimated metal artifact component ($\mathcal{M}$) and the reconstructed image ($\boldsymbol{X}$). Specifically, ResNet is utilized to model the proximal mapping, which helps in projecting the estimated variables onto a more structured solution space, thereby mitigating the effects of noise and artifacts. The effectiveness of CNN-based proximal networks in preserving structural consistency has been further demonstrated by Fu et al.~\cite{fu2024rotation}, highlighting their ability to enforce domain-specific priors and enhance reconstruction fidelity. This choice is particularly beneficial for MAR tasks, as CNN-based architectures leverage spatial correlations and hierarchical feature extraction to improve artifact suppression while maintaining image details. 

FIND-Net extends this approach by integrating frequency-based processing within the iterative framework. Instead of solely relying on spatial domain learning, FIND-Net incorporates frequency-domain priors to refine the artifact estimation and enhance image reconstruction. This hybrid approach allows FIND-Net to capture both local spatial patterns and global frequency characteristics. 


\subsubsection{Network Architecture}
\begin{figure}[h]
    \includegraphics[width=1\linewidth]{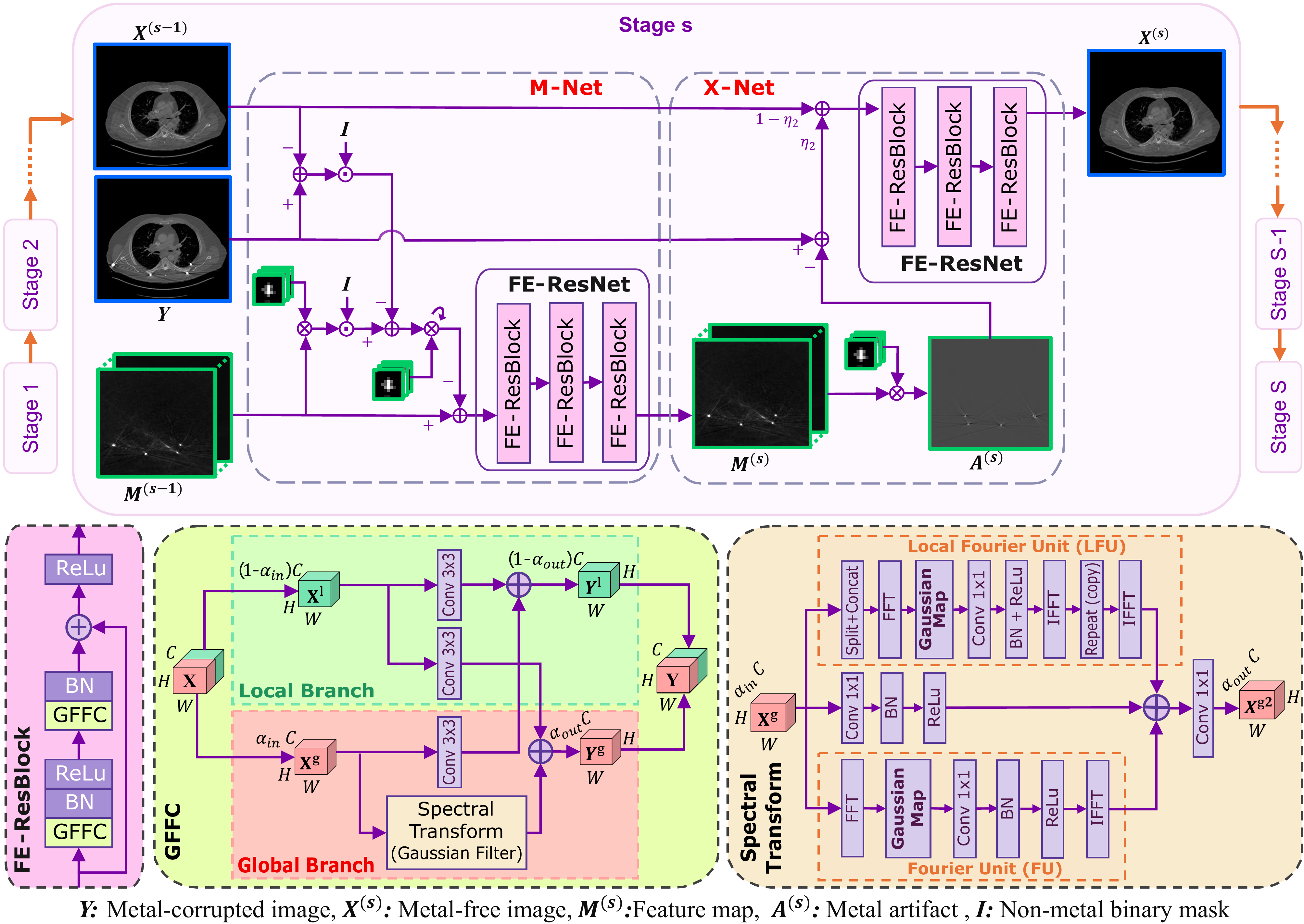}
    \caption{FIND-Net operates in iterative stages: M-Net estimates metal artifacts, while X-Net refines reconstructions. Each stage employs Frequency Enhanced ResNet (FE-ResNet) with Frequency Enhanced ResBlocks (FE-ResBlocks), integrating GFFC, a modified Fast Fourier Convolution (FFC) with trainable Gaussian filtering. GFFC splits channels into Local and Global Branches using \( \alpha \). The Spectral Transform module (LFU, FU) applies Fourier Transform (FFT), Gaussian filtering, and Inverse FFT (IFFT) for frequency enhancement. Symbols: \( \otimes \) (Convolution), \( \otimes^T \) (Transposed Convolution), \( \oplus \) (Element-wise sum), \( \odot \) (Pointwise Multiplication), BN (Batch Norm).}
    \label{fig:FIND-Net_architecture}
\end{figure}

As illustrated in Figure~\ref{fig:FIND-Net_architecture} \footnote{Code is available at \url{https://github.com/Farid-Tasharofi/FIND-Net}}, FIND-Net operates iteratively: M-Net estimates metal artifacts using frequency-enhanced features, while X-Net progressively refines reconstructions by filtering residual artifacts in both spatial and frequency domains. Both networks leverage ResNets for feature extraction. To enhance MAR beyond local feature extraction, FIND-Net replaces standard convolutions in proxNet's ResBlocks with Fast Fourier Convolution (FFC)~\cite{chi2020fast}, enabling joint spatial and frequency-domain processing. FFC partitions input channels into a Local Branch for spatial convolutions and a Global Branch that applies FFT-based spectral transformations, with cross-branch interaction improving feature transfer. The Global Branch employs Fourier Units (FU) and Local Fourier Units (LFU) to capture long-range dependencies while preserving local textures.

To improve frequency adaptability, trainable Gaussian filtering in the Global Branch dynamically adjusts mean and variance during training, selectively enhancing important frequencies while suppressing irrelevant ones. Unlike FFC’s uniform frequency mapping~\cite{chi2020fast}, this adaptive mechanism applies Gaussian filtering before a \(1 \times 1\) convolution in FU and LFU, ensuring robust suppression while preserving anatomical details. The standard FU transformation is:
\[
X^{FU_{out}} = \mathcal{F}^{-1}(\text{ReLU}(\text{BN}(\text{Conv}_{1 \times 1}( \mathcal{F}(X))))).
\]
FIND-Net extends this with trainable Gaussian filtering:
\[
X^{FU_{out}} = \mathcal{F}^{-1}(\text{ReLU}(\text{BN}(\text{Conv}_{1 \times 1}(G_{\sigma,c}(u,v) \cdot  \mathcal{F}(X))))),
\]
where \( G_{\sigma,c}(u,v) =\exp\left(-\left(\frac{D^2 - c^2}{D \cdot \sigma + \epsilon}\right)^2\right) \), with \( D \) as the normalized frequency distance. Learnable \( \sigma \) and \( c \) dynamically control filter bandwidth and center frequency, enabling flexible adaptation to varying artifact patterns.

Channel allocation to the Global Branch increases across stages to balance spatial and spectral learning, starting from \( \alpha_{\text{in}}^{(s)} = \alpha_{\text{out}}^{(s)} = 0.0 \) in the initial stage and reaching \( 0.8 \) in later stages. Early stages emphasize spatial processing, mid-stages incorporate frequency components, and later stages focus on global feature extraction for optimal artifact suppression.

\subsubsection{Training Loss:}

Training employs a multi-term loss at each stage \(s\) to both the restored CT image \( \boldsymbol{X}^{(s)} \) and the extracted artifact \( \boldsymbol{A}^{(s)} \):~\begin{align}
\mathcal{L} = & \sum_{s=0}^S \omega_s \boldsymbol{I} \odot \left\|\boldsymbol{X} - \boldsymbol{X}^{(s)}\right\|_F^2 
+ \gamma_1 \sum_{s=0}^S \omega_s \boldsymbol{I} \odot \left\|\boldsymbol{X} - \boldsymbol{X}^{(s)}\right\|_1 \notag \\
& + \gamma_2 \sum_{s=1}^S \omega_s \boldsymbol{I} \odot \left\|\boldsymbol{Y} - \boldsymbol{X} - \boldsymbol{A}^{(s)}\right\|_1,
\end{align}
where \( \boldsymbol{I} \) is the non-metal mask, and \( \omega_s \) denotes stage-wise weighting. The first term penalizes the Frobenius norm to minimize reconstruction error, while the second and third terms, based on the \( L_1 \) norm, encourage sparsity and enforce consistency between the artifact-free and reconstructed images~\cite{wang2021dicdnet}.

\section{Experiments and Results}
\subsection{Datasets and Experimental Settings}
\subsubsection{Dataset:}
We utilize the AAPM CT-MAR Grand Challenge dataset \cite{AAPM_CT_MAR,AAPM_Benchmark}, generated with XCIST \cite{XCIST} using NIH DeepLesion (lung, abdomen, liver, pelvis) \cite{DeepLesion} and UCLH Stroke EIT (head) \cite{UCLH_Stroke_EIT}, combined with synthetic metal objects. The dataset comprises metal-corrupted and artifact-free CT images, corresponding sinograms, and metal masks, each with a shape of \(512 \times 512\) pixels. For training, 5500 cases were selected (1300 head, 4200 body), with 700 cases each for validation and testing (200 head, 500 body). Realistic metal artifacts were introduced via quantum noise, beam hardening, and scattered radiation. A final evaluation set of 29 cases includes sinograms and metal-corrupted reconstructions but no ground truth, mirroring real-world clinical constraints. This work used the AAPM CT-MAR Grand Challenge datasets \cite{AAPM_CT_MAR,AAPM_Benchmark}. The AAPM CT-MAR Grand Challenge datasets were generated with the open-source CT simulation environment XCIST \cite{XCIST}, using a hybrid data simulation framework that combines publicly available clinical images \cite{DeepLesion,UCLH_Stroke_EIT} and virtual metal objects.

\subsubsection{Experimental Setup and Training Configuration:}
The models were implemented in PyTorch and trained on an NVIDIA A100 GPU (80 GB HBM2) for up to 200 epochs with a batch size of 6. Early stopping was applied based on validation loss. All models trained on the same dataset. DICDNet and FIND-Net were following the configuration set by DICDNet \cite{wang2021dicdnet}, with 10 stages, AdamW (\(\beta_1 = 0.9, \beta_2 = 0.999\)), a learning rate of \(1 \times 10^{-4}\), and weight decay of \(1 \times 10^{-5}\). A linear warmup with cosine annealing was used for learning rate adaptation.

\subsection{Performance Evaluation}
\subsubsection{Quantitative Results:}
We compare FIND-Net against DICDNet, OSCNet, and OSCNet+. OSCNet adds rotation-aware convolutions to DICDNet, while OSCNet+ introduces dynamic convolutions for varying artifact patterns~\cite{wang2023oscnet}. Table~\ref{tab:quant_results} presents synthetic test results across all test cases, showing FIND-Net outperforms all methods in Mean Absolute Error (MAE), Structural Similarity Index (SSIM), and Peak Signal-to-Noise Ratio (PSNR). All deep learning models outperform the traditional LI method. OSCNet variants underperform DICDNet, indicating limited generalization of their rotational symmetry encoding.

\begin{table*}
\centering
\caption{Performance comparison of different MAR approaches in terms of MAE$\downarrow$/SSIM$\uparrow$/PSNR$\uparrow$ across varying metal sizes.}
\label{tab:quant_results}
\renewcommand{\arraystretch}{1.1} 
\fontsize{8pt}{10pt}\selectfont 
\resizebox{\textwidth}{!}{ 
\begin{tabular}{|l|c|c|c|c|}
\hline
Methods & \multicolumn{3}{c|}{Large Metal → Small Metal} & Average \\
\hline
LI & 35.5/0.866/35.9 & 29.6/0.899/38.1 & 23.4/0.929/39.9 & 26.6/0.913/38.9 \\
DICDNet & 23.9/0.918/38.7 & 21.1/0.941/41.2 & 15.6/0.962/43.8 & 18.3/0.951/42.4 \\
OSCNet & 23.2/0.920/39.0 & 21.9/0.938/40.4 & 16.2/0.959/42.9 & 18.8/0.948/41.7 \\
OSCNet+ & 23.5/0.918/39.0 & 21.9/0.938/40.8 & 16.2/0.959/43.2 & 18.8/0.948/42.0 \\
FIND-Net(No-GF) & 24.0/0.917/38.7 & 21.0/0.941/41.3 & 15.5/0.962/43.8 & 18.2/0.951/42.5 \\
\textbf{FIND-Net} & \textbf{22.9}/\textbf{0.925}/\textbf{39.2} & \textbf{20.9}/\textbf{0.942}/\textbf{41.4} & \textbf{15.2}/\textbf{0.963}/\textbf{44.3} & \textbf{17.9}/\textbf{0.952}/\textbf{42.8} \\
\hline
\end{tabular}
}
\end{table*}
\normalsize

\begin{figure}[h]
    \centering
    \includegraphics[width=1\linewidth]{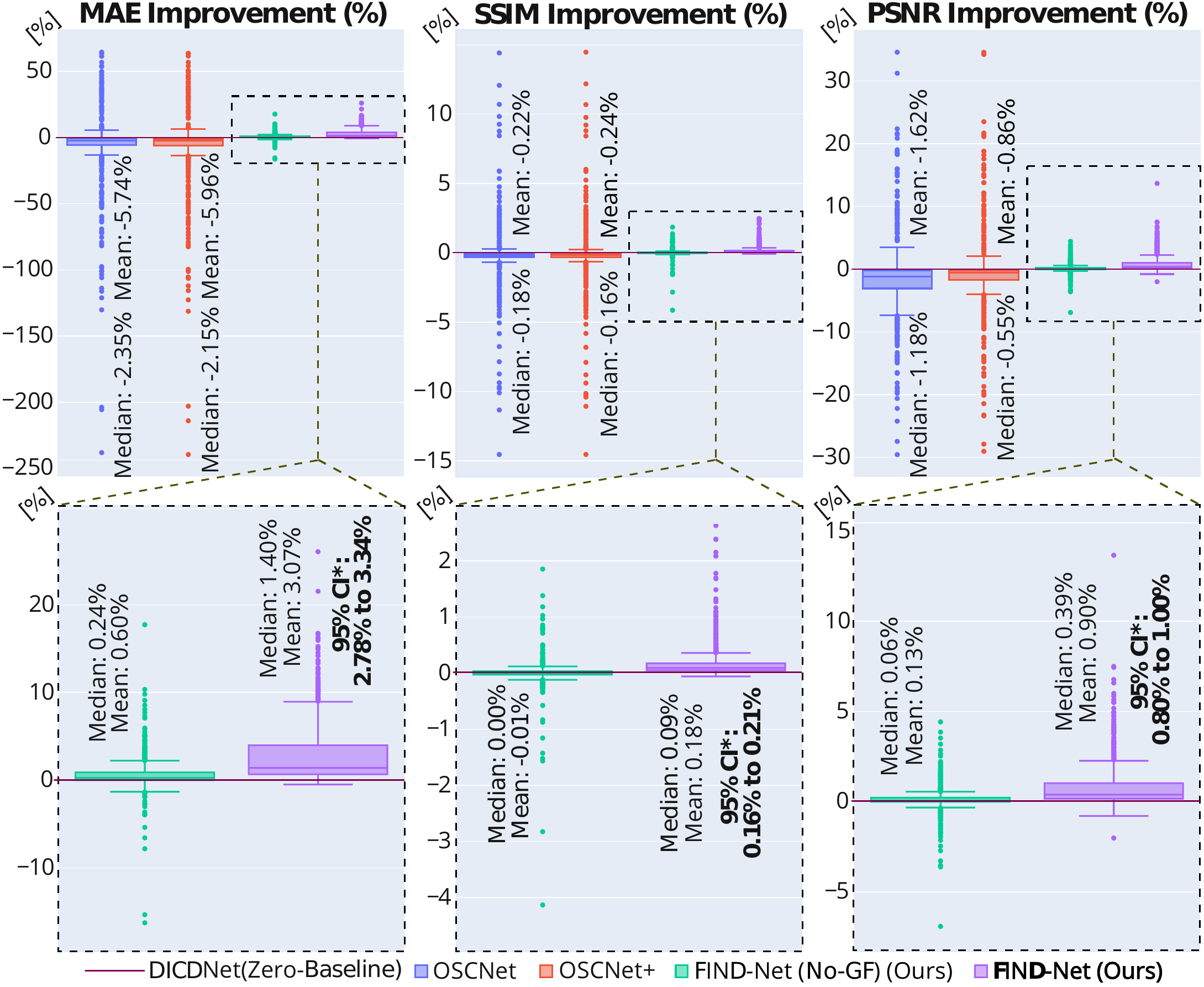}
    \caption{Boxplot comparison of MAE, SSIM, and PSNR improvements (\%) across MAR models, relative to DICDNet. Overall distributions are shown, with zoomed-in views highlighting FIND-Net and FIND-Net without Gaussian filtering (FIND-Net (No-GF)). Mean, median, and 95\% confidence intervals (*) are annotated. A zero-baseline corresponds to DICDNet, and negative values indicate worse performance relative to it.}
    \label{fig:boxplot_improvement}
\end{figure}

We further evaluate FIND-Net’s performance in terms of percentage improvement over the baseline model, DICDNet, as well as OSCNet, OSCNet+, and FIND-Net (No-GF) (FIND-Net without Gaussian Filtering). As illustrated in Figure~\ref{fig:boxplot_improvement}, FIND-Net consistently achieves performance gains, with a 3.07\% reduction in MAE, a 0.18\% increase in SSIM, and a 0.90\% improvement in PSNR compared to DICDNet. The annotated 95\% confidence intervals confirm the statistical significance of these improvements. The boxplots further highlight FIND-Net’s robust performance, with the zoomed-in views emphasizing the contribution of Gaussian Filtering in enhancing frequency-domain learning. Notably, the reduction in MAE indicates superior artifact suppression, reinforcing FIND-Net’s effectiveness in metal artifact reduction.

\subsubsection{Computational Complexity:}
FIND-Net reduces computational complexity by approximately \textbf{16\%} (334.45 GFLOPs → 279.55 GFLOPs) by leveraging FFT-based optimizations. However, inference time increases from 0.15s to 1.07s per image due to the overhead of frequency-domain transformations. 

\subsubsection{Qualitative Results}

\begin{figure}[t]
    \centering
    \includegraphics[width=1\linewidth]{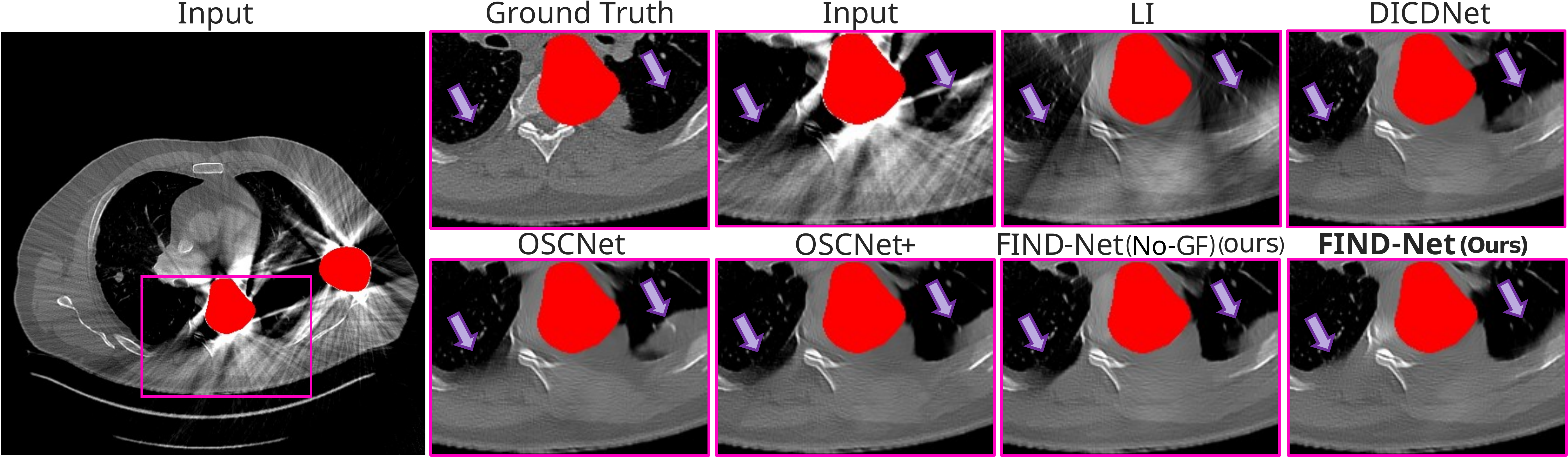}
    \caption{Qualitative comparison of MAR results on a metal-corrupted synthetic CT image. The highlighted ROIs demonstrate FIND-Net’s superior artifact suppression, particularly in fine streak removal and edge preservation, compared to other methods. The red mask highlights metal mask. (No-GF stands for no Gaussian filtering)}

    \label{fig:qualitative_comparison}
\end{figure}

Figure~\ref{fig:qualitative_comparison} compares FIND-Net to other models on a synthetic test case. While LI partially suppresses streaks, DICDNet improves artifact reduction, particularly in the right-arrow region, outperforming OSCNet variants. FIND-Net further refines reconstruction, as Gaussian filtering eliminates low-contrast artifacts persisting in other models. The left-arrow region highlights FIND-Net’s ability to enhance edge sharpness and reduce excessive smoothing, preserving anatomical boundaries. This ensures superior preservation and artifact suppression in both high- and low-contrast areas. 

\begin{figure}[h]
    \centering
    \includegraphics[width=1.0\linewidth]{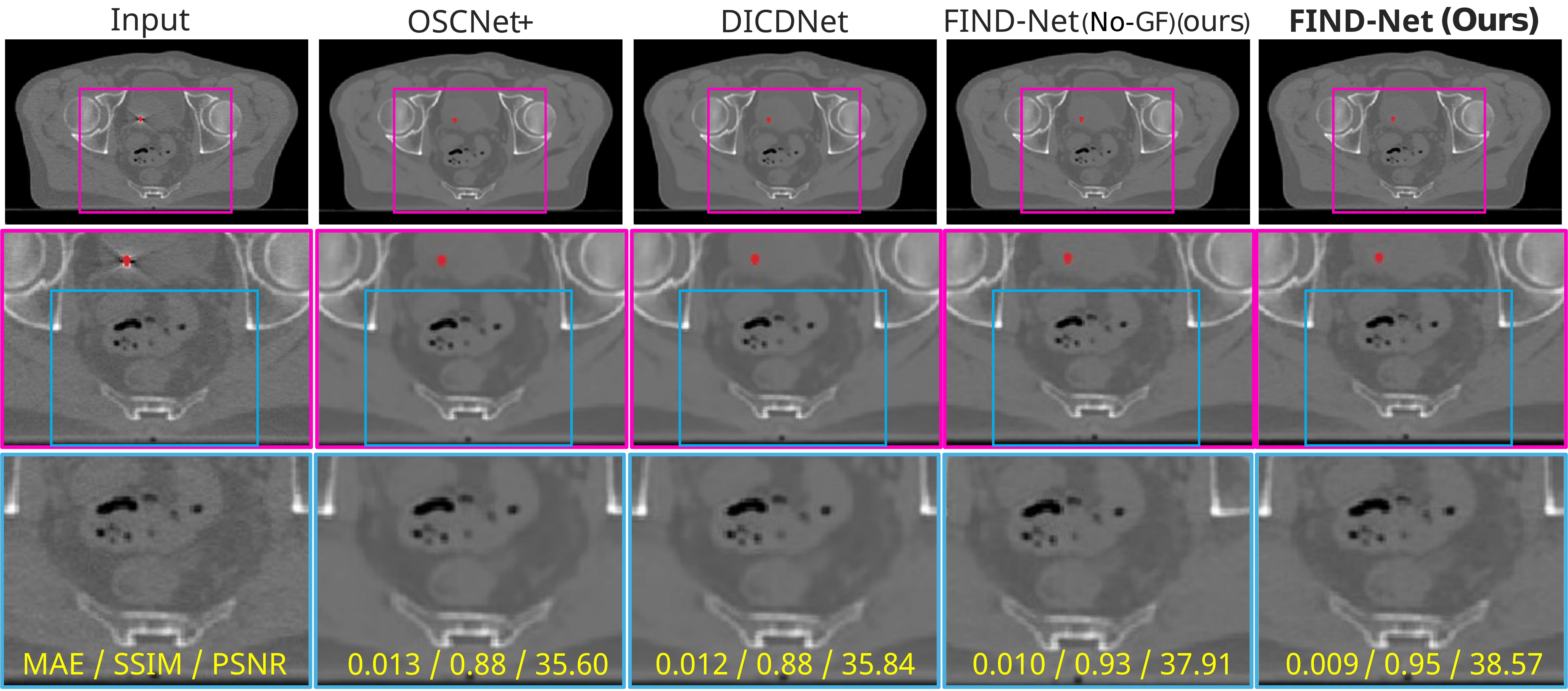}
    \caption{Qualitative comparison of MAR methods on a real-world CT scan (no ground truth available). The red mask highlights segmented metal. The blue region (artifact-free in the input) is evaluated for structural preservation. }

    \label{fig:real_world_comparison}
\end{figure}

Figure~\ref{fig:real_world_comparison} shows results on a real-world CT scan from our 29-image dataset, where no artifact-free reference is available. While FIND-Net and DICDNet reduce artifacts, DICDNet and OSCNet variants introduce excessive smoothing in non-corrupted regions. The blue region analysis confirms FIND-Net’s superior preservation, achieving the lowest MAE and highest SSIM and PSNR. 

To evaluate the impact on clean anatomical regions, we analyzed artifact-free patches from seven images. An ideal MAR method should leave these patches unchanged. OSCNet, OSCNet+, and DICDNet perform similarly (MAE $\approx$ 0.007, SSIM $\approx$ 0.94, PSNR $\approx$ 40.4), while FIND-Net without Gaussian filtering better preserves structures (MAE 0.006, SSIM 0.96, PSNR 41.81). FIND-Net outperforms all models, achieving the best scores (MAE 0.005, SSIM 0.97, PSNR 43.23) and demonstrating robust suppression with minimal unintended modifications.

\section{Conclusion}
This paper introduces FIND-Net, a deep learning-based MAR framework integrating FFC and trainable Gaussian filtering for enhanced artifact suppression. By combining spatial and frequency-domain processing, FIND-Net outperforms existing MAR methods, achieving a statistically significant 3.07\% MAE reduction, 0.18\% SSIM increase, and 0.90\% PSNR improvement over DICDNet. Comprehensive experiments confirm FIND-Net’s robustness across varying artifact complexities, demonstrating superior suppression of streak artifacts while preserving anatomical structures. Despite an increased inference time (1.07s per image) due to frequency-domain operations, FIND-Net reduces computational complexity and sets a new benchmark for MAR performance.

    

\begin{credits}
\subsubsection{\ackname} The authors gratefully acknowledge the scientific support and HPC resources provided by the Erlangen National High Performance Computing Center of the Friedrich-Alexander-Universität Erlangen-Nürnberg. The hardware is funded by the German Research Foundation. This research has received funding from the scholarship program “Dhip campus-bavarian aim”.

\subsubsection{\discintname}
The authors have no competing interests to declare that are relevant to the content of this article.
\end{credits}

%
%
%

\begin{thebibliography}{8}
\bibitem{gjesteby2016metal}
Gjesteby, L., De Man, B., Jin, Y., Paganetti, H., Verburg, J., Giantsoudi, D., Wang, G.: Metal artifact reduction in CT: where are we after four decades? IEEE Access \textbf{4}, 5826--5849 (2016). \doi{10.1109/ACCESS.2016.2608621}

\bibitem{wang2023oscnet}
Wang, H., Xie, Q., Zeng, D., Ma, J., Meng, D., Zheng, Y.: OSCNet: Orientation-Shared Convolutional Network for CT Metal Artifact Learning. IEEE Trans. Med. Imaging \textbf{43}(1), 489--502 (2024). \doi{10.1109/TMI.2023.3310987}


\bibitem{anhaus2024new}
Anhaus, J., Hofmann, C.: New Approaches for Metal Artifact Reduction in Computed Tomography. In: Iniewski, K., Gadey, H. (eds.) Emerging Radiation Detection: Technology and Applications, pp. 131--146. Springer, Cham (2024). \doi{10.1007/978-3-031-63897-8_8}

\bibitem{kalender1987reduction}
Kalender, W.A., Hebel, R., Ebersberger, J.: Reduction of CT artifacts caused by metallic implants. Radiology \textbf{164}(2), 576--577 (1987). \doi{10.1148/radiology.164.2.3602406}


\bibitem{Lyu_2020}
Lyu, Y., Lin, W.A., Liao, H., Lu, J., Zhou, S.K.: Encoding Metal Mask Projection for Metal Artifact Reduction in Computed Tomography. In: Martel, A.L., et al. (eds.) MICCAI 2020. LNCS, vol. 12262, pp. 147--157. Springer, Cham (2020). \doi{10.1007/978-3-030-59713-9_15}

\bibitem{yu2020deep}
Yu, L., Zhang, Z., Li, X., Xing, L.: Deep sinogram completion with image prior for metal artifact reduction in CT images. IEEE Trans. Med. Imaging \textbf{40}(1), 228--238 (2021). \doi{10.1109/TMI.2020.3025064}


\bibitem{wang2021dicdnet}
Wang, H., Li, Y., He, N., Ma, K., Meng, D., Zheng, Y.: DICDNet: deep interpretable convolutional dictionary network for metal artifact reduction in CT images. IEEE Trans. Med. Imaging \textbf{41}(4), 869--880 (2022). \doi{10.1109/TMI.2021.3127074}


\bibitem{richter2021should}
Richter, M.L., Schöning, J., Wiedenroth, A., Krumnack, U.: Should you go deeper? Optimizing convolutional neural network architectures without training. In: 20th IEEE International Conference on Machine Learning and Applications (ICMLA), pp. 964--971. IEEE, Pasadena, CA, USA (2021). \doi{10.1109/ICMLA52953.2021.00159}



\bibitem{chi2020fast}
Chi, L., Jiang, B., Mu, Y.: Fast Fourier Convolution. In: Larochelle, H., Ranzato, M., Hadsell, R., Balcan, M.F., Lin, H. (eds.) Advances in Neural Information Processing Systems 33 (NeurIPS 2020), pp. 4479--4488. Curran Associates, Inc. (2020). 

\bibitem{he2016deep}
He, K., Zhang, X., Ren, S., Sun, J.: Deep residual learning for image recognition. In: Proceedings of the IEEE Conference on Computer Vision and Pattern Recognition (CVPR), pp. 770--778 (2016). \doi{10.48550/arXiv.1512.03385}

\bibitem{fu2024rotation}
Fu, J., Xie, Q., Meng, D., Xu, Z.: Rotation equivariant proximal operator for deep unfolding methods in image restoration. IEEE Trans. Pattern Anal. Mach. Intell. \textbf{46}(10), 6577--6593 (2024). \doi{10.1109/TPAMI.2024.3383532}


\bibitem{AAPM_CT_MAR}
AAPM CT Metal Artifact Reduction (CT-MAR) Grand Challenge. \url{https://www.aapm.org/GrandChallenge/CT-MAR/}, last accessed 2025/02/01.


\bibitem{AAPM_Benchmark}
AAPM CT Metal Artifact Reduction (CT-MAR) Grand Challenge Benchmark Tool, \url{https://github.com/xcist/example/tree/main/AAPM\_datachallenge},  last accessed 2025/02/01



\bibitem{XCIST}
Wu, M., FitzGerald, P., Zhang, J., Segars, W.P., Yu, J., Xu, Y., De Man, B.: XCIST - an open access x-ray/CT simulation toolkit. Phys Med Biol. \textbf{67}(19) (2022). \doi{10.1088/1361-6560/ac8c85}



\bibitem{DeepLesion}
Yan, K., Wang, X., Lu, L., Summers, R.M.: DeepLesion: Automated Mining of Large-Scale Lesion Annotations and Universal Lesion Detection with Deep Learning. J Med Imaging. \textbf{5}(3), 036501 (2018). \doi{10.1117/1.JMI.5.3.036501}

\bibitem{UCLH_Stroke_EIT}
Goren, N., Dowrick, T., Avery, J., Holder, D.: UCLH Stroke EIT Dataset - Radiology Data (CT). Zenodo (2017). \doi{10.5281/zenodo.838704}


\end{thebibliography}
%

\end{document}